\documentclass[sigconf]{acmart}

\renewcommand\footnotetextcopyrightpermission[1]{} 
\usepackage[ruled,vlined]{algorithm2e}
\usepackage{subcaption}
\usepackage[font=small,labelfont=bf]{caption}

\begin{document}

\title{Anomaly Detection for High-Dimensional Data Using Large Deviations Principle}

\author{Sreelekha Guggilam}
\email{sreelekh@buffalo.edu}
\affiliation{%
  \institution{University at Buffalo}
  \city{Buffalo}
  \state{NY}
  \country{USA}
}
\author{Varun Chandola}
\email{chandola@buffalo.edu}
\affiliation{%
  \institution{University at Buffalo}
  \city{Buffalo}
  \state{NY}
  \country{USA}
}
\author{Abani Patra}
\email{abani.patra@tufts.edu}
\affiliation{%
  \institution{Tufts University}
  \city{Boston}
  \state{MA}
  \country{USA}
}

\begin{abstract}
Most current anomaly detection methods suffer from the {\em curse of dimensionality} when dealing with high-dimensional data. We propose an anomaly detection algorithm that can scale to high-dimensional data using concepts from the theory of {\em large deviations}. The proposed {\em Large Deviations Anomaly Detection} (LAD) algorithm is shown to outperform state of art anomaly detection methods on a variety of large and high-dimensional benchmark data sets. Exploiting the ability of the algorithm to scale to high-dimensional data, we propose an online anomaly detection method to identify anomalies in a collection of multivariate time series. We demonstrate the applicability of the online algorithm in identifying counties in the United States with anomalous trends in terms of COVID-19 related cases and deaths. Several of the identified anomalous counties correlate with counties with documented poor response to the COVID pandemic.
\end{abstract}





\maketitle
\pagestyle{plain}

\section{Introduction}
Anomaly detection has been extensively studied over many decades across many domains~\cite{chandola2009anomaly, Hodge:2004}.  Among the most useful applications of anomaly detection is to simultaneously monitor multiple systems' behaviors and identify the system  that exhibits anomalous behavior due to external or internal stress factors. For instance, consider the example of the COVID-19 infection data. Studying the confirmed case and death trends across various countries, states or counties could highlight and identify the most (or least) significant public policies. One possible approach to study the data could be to monitor each time series~\cite{ceylan2020estimation, maleki2020time, zeroual2020deep} and identify sudden outbreaks or significant causal events. However, such methods study each time series individually and cannot not be used to detect the gradual divergence from the normal trends or initial signs of such drift.

An alternate approach is to analyze each time series in the context of a collection of time series, which can reveal anomalies beyond sudden and significant events, such as anomalous trends and gradual drifts. Such methods typically require an appropriate similarity measure~\cite{fu2011review}. Through appropriate combination with state-of-the-art similarity-based models, these methods can identify potential anomalous time series and cluster similar trends. Implementing such methods in a time varying setting could even help detect change points or anomalous events in individual time series as well as identifying anomalous time series~\cite{zhang2019deep, benkabou2018unsupervised}. However, these methods are typically unable to scale to long time series~\cite{zhang2019deep, beggel2019time}.

\begin{figure*}
     \centering
     \begin{subfigure}[b]{0.78\textwidth}
         \centering
         \includegraphics[width=\textwidth]{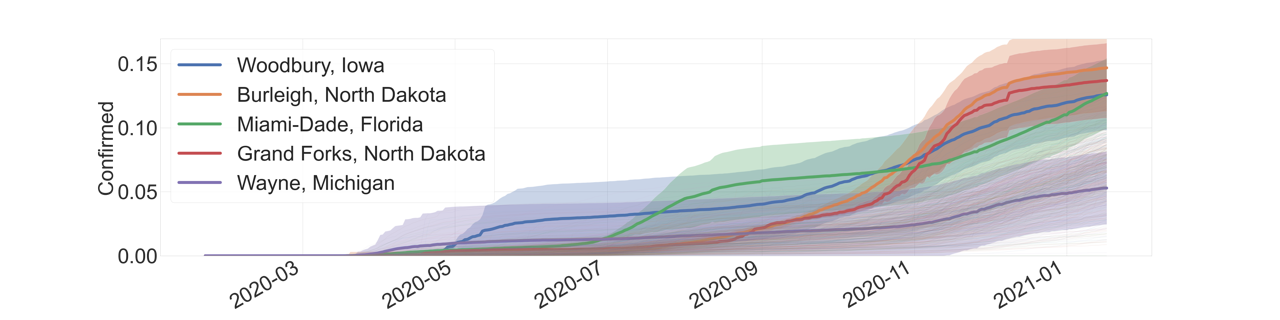}
         \caption{Total Confirmed Cases}
     \label{fig:my_label}
     \end{subfigure}
     \begin{subfigure}[b]{0.78\textwidth}
         \centering
         \includegraphics[width=\textwidth]{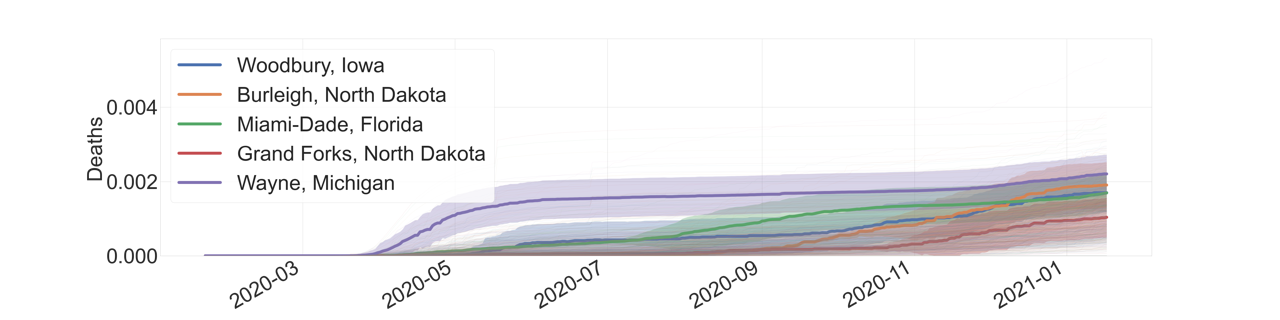}
         \caption{Total Deaths}
     \label{fig:d_total_full_hist_combined_intro}
     \end{subfigure}
     \caption{Top 5 anomalous counties identified by the proposed LAD algorithm based on the daily multivariate time-series, consisting of cumulative COVID-19 per-capita infections and deaths. At any time-instance, the algorithm analyzes the bi-variate time series for all the counties to identify anomalies. The time-series for the non-anomalous counties are plotted (light-gray) in the background for reference. For the counties in North Dakota (Burleigh and Grand Forks), the number of confirmed cases ({\em top}), and the sharp rise in November 2020, is the primary cause for anomaly\protect\footnotemark. On the other hand, Wayne County in Michigan was identified as anomalous primarily because of its abnormally high death rate, especially when compared to the relatively moderate confirmed infection rate.} 
     \label{fig:my_labelf}
\end{figure*}
\footnotetext{In early November, these counties in North Dakota were exhibiting infection rates that were six times the national rate - \url{https://www.washingtonpost.com/opinions/2020/11/06/north-dakota-covid-19-cases/}}


In this paper, we propose a new anomaly detection algorithm called {\em Large deviations Anomaly Detection} (LAD), for large/high-dimensional data and multivariate time series data. LAD uses the rate function from {\em large deviations principle} (LDP)~\cite{den2008large, varadhan1984large, varadhan2010large} to deduce anomaly scores for the underlying data. Core ideas for the algorithm are inspired from large deviation theory's projection theorem that allow better handling of high dimensional data. Unlike most high dimensional anomaly detection models, LAD does not incorporate feature selection or dimensionality reduction, which makes it ideal to study multiple time series in an online mode. The intuition behind the LAD model allows it to naturally segregate the anomalous observations at each time step while comparing multiple multivariate time series simultaneously. The key contributions of this paper are following:
\begin{enumerate}
    \item We propose the {\em Large deviations Anomaly Detection} (LAD) algorithm, a novel and highly scalable LDP based methodology, for scoring based anomaly detection. 
    \item The proposed LAD model is capable of analyzing large and high dimensional datasets without additional dimensionality reduction procedures thereby allowing more accurate and cost effective 
    anomaly detection.
    \item An online extension of the LAD model is presented  to detect anomalies in an multivariate time series database using an evolving anomaly score for each time series. The anomaly score varies with time and can be used to track developing anomalous behavior.
    \item We perform an empirical study on publicly available anomaly detection benchmark datasets to analyze robustness and performance of the proposed method on high dimensional and large datasets.
    \item We present a detailed analysis of COVID-19 trends for US counties where we identify counties with anomalous behavior (See Figure~\ref{fig:my_labelf} for an illustration).
\end{enumerate}

The rest of this document is organized as follows. Section \ref{lit_review} provides an overview of relevant existing methods for anomaly detection. Section \ref{ldp_background} is a short background on underlying large deviations theory motivating LAD. Section \ref{methodology} details our LAD model for detecting unsupervised anomalies in multivariate time series. Section \ref{experiments} describes the experiments and demonstrate the state-of-the-art performance of our method. Section \ref{conclusion} concludes the paper and sketches direction for possible future work.

\section{Related Work}\label{lit_review}
In this section, we provide a brief overview of relevant anomaly detection methods which have been proposed for high-dimensional data and for multivariate time-series data. We also discuss other works that have used the large deviations principle for detecting anomalies.

A large body of research exists on studying anomalies in high dimensional data \cite{aggarwal2001outlier, angiulli2002fast} but challenges remain. Many anomaly detection algorithms use dimensionality reduction techniques as a pre-processing step to anomaly detection. However, many high dimensional anomalies can only be detected in high dimensional problem settings and dimensionality reduction in such settings can lead to false negatives. Many methods exist that identify anomalies on high-dimensional data without dimensional reduction or feature selection, e.g. by using distance metrics. {\em Elliptic Envelope} (EE)~\cite{rousseeuw1999fast} fits an ellipse around data centers by fitting a robust covariance estimates.  {\em Isolation Forest} (I-Forest)~\cite{liu2012isolation} uses recursive partitioning by random feature selection and isolating outlier observations. {\em $k$ nearest neighbor outlier detection} (kNN)~\cite{Ramaswamy:2000} uses distance from nearest neighbor to get anomaly scores. {\em local outlier factor} (LOF)~\cite{Breunig:2000} uses deviation in local densities with respect to its neighbors to detect anomalies. {\em k-means\texttt{-{}-}}~\citep{chawla2013k} method uses distance from nearest cluster centers to jointly perform clustering and anomaly detection. {\em Concentration Free Outlier Factor} (CFOF) \cite{angiulli2020cfof} uses a ``reverse nearest neighbor-based score'' which measures the number of nearest neighbors required for a point to have a set proportion of data within its envelope. In particular, methods like I-Forest and CFOF are targeted towards anomaly detection in high dimensional datasets. 

In most settings, real time detection of anomalies is needed to dispatch necessary preventive measures for damage control. Such problem formulation requires collectively monitoring a high dimensional time series database to identify anomalies in real time. Recently, large deviations theory has been widely applied in the fields of climate models \cite{dematteis2018rogue}, statistical mechanics \cite{touchette2009large}, networks \cite{ paschalidis2008spatio}, etc. Specially for analysis of time series, the theory of large deviations has proven to be of great interest over recent decades \cite{boettiger2013no, mikosch2016large}. However, these methods are data specific, often study individual time series and are difficult to generalize to other areas of research.

Anomaly detection for time series have been extensively explored in the literature~\cite{gupta2013outlier}, though most focus has been on identifying anomalous events in a single time-series. While, the task of detecting anomalous time series in a collection of time series has been studied in the past~\cite{yankov2008disk, Chandola:2008, Chandola:2009a}, most of these works have focused on univariate time series and have not shown to scale to long time series data. Our proposed method addresses this issue by using the large deviation principle.

\section{Large Deviation Principle}\label{ldp_background}
Large deviations theory provides techniques to derive the probability of rare events\footnote{In our context, these rare events include outlier/anomalous behaviors.} 
that have an asymptotically exact exponential approximation\cite{den2008large, varadhan1984large, varadhan2010large}. 
In this section, we briefly go over the large deviation theory and different ways to generate the rate functions required for the large deviations principle.

The key concept of this theory is the Large Deviations Principle (LDP). The principle describes the exponential decay of the probabilities for the mean of random variables. The rate of decay is characterized by the rate function $\mathcal{I}$. The theorem is detailed below:

\begin{theorem}\label{ldp_theorem}
    A family of probability measures $\{\mu_\epsilon\}_{\epsilon>0}$ on a Polish space $\mathcal{X}$ is said to satisfy large deviation principle (LDP) with the rate function $\mathcal{I}:\mathcal{X}\to[0,\infty]$ if:
    \begin{enumerate}
        \item $\mathcal{I}$ has compact level sets and is not identically infinite
        \item $lim inf_{\epsilon \to 0} \epsilon log\mu_\epsilon(\mathcal{O})\geq-\mathcal{I}(\mathcal{O}) \quad \forall \mathcal{O}\subseteq \mathcal{X}$ open sets
        \item $lim sup_{\epsilon\to0} \epsilon log\mu_\epsilon(\mathcal{C}) \leq-\mathcal{I}(\mathcal{C}) \quad \forall \mathcal{C}\subseteq \mathcal{X}$ closed sets
    \end{enumerate}
where, $\mathcal{I}(\mathcal{S})=inf_{x\in \mathcal{S}}\mathcal{I}(x),\ \mathcal{S}\subseteq\mathcal{X}$ 
\end{theorem}

To implement LDP on known data with known distributions, it is important to decipher the rate function $\mathcal{I}$. Cramer's Theorem provides the relation between the rate function $\mathcal{I}$ and the logarithmic moment generating function $\Lambda$. 

\begin{definition}
The logarithmic moment generating function of a random variable $X$ is defined as
\begin{equation}
    \begin{aligned}
        \Lambda(t)=\log E[\exp(tX)]
    \end{aligned}
\end{equation}
\end{definition}

\begin{theorem}[Cramer's Theorem]
Let $ X_1, X_2, \dots X_n$ be a sequence of iid real random variables with finite logarithmic moment generating function, e.g. $\Lambda(t)<\infty$ for all $t\in\mathbb{R}$.
Then the law for the empirical average satisfies the large deviations principle with rate $\epsilon=1/n$ and rate function given by
\begin{equation}
    \begin{aligned}
        \mathcal{I}(x):= \sup_{t \in \mathbb{R}} \left(tx-\Lambda(t) \right) \quad \forall t\in\mathbb{R}
    \end{aligned}
\end{equation}
\end{theorem}
Thus, we get, 
\begin{equation}
    \begin{aligned}
    \lim_{n \to \infty} \frac{1}{n} \log \left(P\left(\sum_{i=1}^n X_i \geq nx \right)\right) = -\mathcal{I}(x),\quad \forall x > E[X_1]
    \end{aligned}
\end{equation}


For more complex distributions, identifying the rate function using logarithmic moment generating function can be challenging. Many methods like contraction principle and exponential tilting exist that extend rate functions from one topological space that satisfies LDP to the topological spaces of interest\cite{den2008large}. For our work, we are interested in the Dawson-G\"artner Projective LDP, that generates the rate function using nested family of projections. 

\begin{theorem}{Dawson-G\"artner Projective LDP: }
Let $\{\pi^N\}_{N\in\mathbb{N}}$ be a nested family of projections acting on $\mathcal{X}$ s.t. $\cup_{N\in\mathbb{N}}\pi^N$ is the identity. Let $\mathcal{X}^N=\pi^N\mathcal{X}$ and $\mu_\epsilon^N=\mu_0\circ(\pi^N)^{-1}, N\in\mathbb{N}$. If $\forall N\in\mathcal{N}$, the family $\{\mu^N_\epsilon\}_{\epsilon>0}$ satisfies the LDP on $\mathcal{X}^N$ with rate function $\mathcal{I}^N$, then $\{\mu_\epsilon\}_{\epsilon>0}$ satisfies the LDP with rate function $I$ given by,
\[
\begin{aligned}
\mathcal{I}(x)=sup_{N\in\mathbb{N}}\mathcal{I}^N(\pi^N x)\quad x\in\mathcal{X}
\end{aligned}
\]
Since $\mathcal{I}^N(y)=inf_{\{ x\in\mathcal{X}\vert \pi^N(x)=y\}}\mathcal{I}(x),\ y\in\mathcal{Y}$, the supremum defining $\mathcal{I}$ is monotone in N because projections are nested.
\end{theorem}

The theorem allows extending the rate function from a lower projection to higher projection space. The implementation of this theorem in LAD model is discussed in Section \ref{methodology}.



\section{Methodology}\label{methodology}
Consider the case of multivariate time series data. 
Let $\{\bf{t}_n\}_{n=1}^{N}$ be a set of multivariate time series datasets where $\bf{t}_n={(\bf{t}_{n,1},\dots,\bf{t}_{n,T})}$ is a time series of length $T$ and each $\bf{t}_{n,t}$ has $d$ attributes. The motivation is to identify anomalous $\bf{t}_n$ that diverge significantly from the non-anomalous counter parts at a given or multiple time steps. 

The main challenge is to design a score for individual time series that evolves in a temporal setting as well as enables tracking the initial time of deviation as well as the scale of deviation from the normal trend. 

As shown in following sections,  our model addresses the problem through the use of rate functions derived from large deviations principle. We use the Dawson-G\"artner Projective LDP (See Section~\ref{ldp_hd}) for projecting the rate function function to a low dimensional setting while preserving anomalous instances. 

The extension to temporal data (See Section~\ref{ldp_ts}) is done by collectively studying each time series data as one observation. 


\subsection{Large Deviations for Anomaly Detection}
Our approach uses a direct implementation of LDP to derive the rate function values for each observation. As the theory focuses on extremely rare events, the raw probabilities associated with them are usually very small \cite{varadhan1984large, den2008large, varadhan2010large}. However, the LDP provides a rate function that is useful as a scoring metric for our LAD model. 

Consider a dataset $X$ of size $n$. Let $\bf{a}=\{a_1,\dots,a_n\}$ and $\bf{I}=\{I_1,\dots,I_n\}$ be anomaly score and anomaly label vectors for the observations respectively such that $a_i\in[0,1]$ and $I_i\in\{0,1\}$ $\forall i \in \{1,2,\dots,n\}$. 

By large deviations principle, we know that for a given dataset $X$ of size $n$, $P(\bar{X}=p) \approx e^{-n\mathcal{I}(p)}$. Assuming that the underlying data is standard Gaussian distribution with mean 0 and variance 1, we can use the rate function for Gaussian data where $\mathcal{I}(p)=\frac{p^2}{2}$. Then the resulting probability that the sample mean is $p$ is given by:
\begin{equation}
    \begin{aligned}
    P(\bar{X}=p) \approx e^{-n\frac{p^2}{2}}
    \end{aligned}
\end{equation}

Now, in presence of an anomalous observation $x_a$, the sample mean is shifted by approximately $x_a/n$ for large $n$. Thus, the probability of the shifted mean being the true mean is given by,
\begin{equation}
    \begin{aligned}
        P(\bar{X}=x_a/n) \approx e^{-\frac{x_a^2}{2n}}
    \end{aligned}
\end{equation}

However, for large n and $\vert x_a\vert <<1$, the above probabilities decay exponentially which significantly reduces their effectiveness for anomaly detection. Thus, we use $\frac{x_a^2}{2n}$ as anomaly score for our model.
Thus generalizing this, the anomaly score for each individual observation is given by:
\begin{equation}
    \begin{aligned}
        a_i=n\mathcal{I}(x_i)\quad\forall i \in \{1,2,\dots,n\}
    \end{aligned}\label{ldp_score}
\end{equation}

\subsection{LDP for High Dimensional Data}\label{ldp_hd}
High dimensional data pose significant challenges to anomaly detection. Presence of redundant or irrelevant features act as noise making anomaly detection difficult. However, dimensionality reduction can impact anomalies that arise from less significant features of the datasets. To address this, we use the Dawson-G\"artner Projective theorem in LAD model to compute the rate function for high dimensional data. The theorem records the maximum value across all projections which preserves the anomaly score making it optimal to detect anomalies in high dimensional data. The model algorithm is presented in Algorithm \ref{alg:generator1}. 

\begin{algorithm}
\caption{Algorithm 1: LAD Model}
\label{alg:generator1}
\SetKwProg{generate}{Function \emph{generate}}{}{end}
\textbf{Input}: Dataset $X$ of size $(n,d)$, number of iterations $N_{iter}$, threshold $th$. \\
\textbf{Output}: Anomaly score $\bf{a}$ \\
\textbf{Initialization}: Set initial anomaly score and labels $\bf{a}$ and $\bf{I}$ to zero vectors and, entropy matrix $E=0_{(n,d)}$ where $0_{(n,d)}$ is a zero matrix of size $(n,d)$.\\
\For{each  $s\rightarrow 1\ \KwTo\ N_{iter}$}{
    \begin{enumerate}
        \item Subset $X_{sub}=X[{I}_i==0]$
        \item $X_{normalized}[:,d_i]=\frac{X[:,d_i]-\bar{X_{sub}[:,d_i]}}{cov(X_{sub}[:,d_i])}, \quad \forall d_i\in\{1,\dots,d\}$
        \item $E[i,:]=-X_{normalized}[i]^{2}/2n, \quad \forall i$
        \item $a_i=-max(E[i,:])$
        \item $\bf{a}=\frac{\bf{a}-min(\bf{a})}{max(\bf{a})-min(\bf{a})}$
        \item $th=min(th,quantile(\bf{a},0.95)$
        \item $I_i=1$ if $a_i>th,  \quad \forall i$
     \end{enumerate}}
\end{algorithm}

\subsection{LAD for Time Series Data}\label{ldp_ts}
The definition of an anomaly is often contingent on the data and the problem statement. Broadly, time series anomalies can be categorized to two groups \cite{Chandola:2009a}:
\begin{enumerate}
    \item \textbf{Divergent trends/Process anomalies}: Time series with divergent trends that last for significant time periods fall into this group. Here, one can argue that generative process of such time series could be different from the rest of the non-anomalous counterparts.   
    \item \textbf{Subsequence anomalies}: Such time series have temporally sudden fluctuations or deviations from expected behavior which can be deemed as anomalous. These anomalies occur as a subsequence of sudden spikes or fatigues in a time series of relatively non-anomalous trend. 
\end{enumerate}

The online extension of the LAD model is designed to capture anomalous behavior at each time step. Based on the mode of analysis of the temporal anomaly scores, one can identify both divergent trends and subsequence anomalies. In this paper, we focus on the divergent trends (or process anomalies). In particular, we try to look at the anomalous trends in COVID-19 cases and deaths in US counties. Studies to collectively identify divergent trends and subsequence anomalies is being considered as a prospective future work. 

In this section, we present an extension of the LAD model to multivariate time series data. Here, we wish to preserve the temporal dependency as well as dependency across different features of the time series. Thus, as shown in Algorithm \ref{alg:generator2}, a horizontal stacking of the data is performed. This allows collective study of temporal and non-temporal features. To preserve temporal dependency, the anomaly scores and labels are carried on to next time step where the labels are then re-evaluated. 

\begin{algorithm}
\caption{Algorithm 2: LAD for Time series anomaly detection}
\label{alg:generator2}
\SetKwProg{generate}{Function \emph{generate}}{}{end}
\textbf{Input}: Time series dataset $\{\bf{t}_n\}_{n=1}^{N}$ of size $(N,T,d)$, number of iterations $N_{iter}$, threshold $th$, window $w$.\\
\textbf{Output}: An array of temporal anomaly scores $\bf{a}$, an array of temporal anomaly labels $I$ \\
\textbf{Initialization}: Set initial anomaly score and labels $\bf{a}$ and $\bf{I}$ to zero matrices of size $(N,T)$ and, entropy matrix $E$ to a zero matrix of size $(N,T,d)$.\\
\For{each  $t\rightarrow 1\ \KwTo\ T$}{
$X=hstack(\bar{t_{n,t}})$ where $\bar{t_{n,t}}=\{t_{n,t-w},\dots t_{n,t}\}$\\
$I[i,t]=I[i,t-1]$\\
$\bf{a}[:,t]=\bf{a}[:,t-1]$\\
\For{each  $s\rightarrow 1\ \KwTo\ N_{iter}$}{
    \begin{enumerate}
        \item Subset non-anomalous time series $X_{sub}=\{X[i,:] \vert I[i,t]==0, \forall i\}$
        \item $X_{normalized}[:,d_i]=\frac{X[:,d_i]-\bar{X_{sub}[:,d_i]}}{cov(X_{sub}[:,d_i])}, \quad \forall d_i \in \{1,2,\dots,d*w\}$ 
        \item $E[i,:]=-X_{normalized}[i]^{2}/2n, \quad \forall i$
        \item $\bf{a}[i,t]=-max(E[i,:])$
        \item $\bf{a}[:,t]=\frac{\bf{a}[:,t]-min(\bf{a}[:,t])}{max(\bf{a}[:,t])-min(\bf{a}[:,t])}$
        \item $th=min(th,quantile(\bf{a}[:,t],0.95)$
        \item $I[i,t]=1$ if $\bf{a}[i,t]>th,  \quad \forall i$
     \end{enumerate}}}
\end{algorithm}

As long term anomalies are of interest, time series with temporally longer anomalous behaviors are ranked more anomalous. The overall time series anomaly score $A_n$ for each time series $\bf{t}_n$ can be computed as:
\begin{equation}
    \begin{aligned}
        A_n=\frac{\sum_{t=1}^{T} I[n,t]}{T} \quad \forall n
    \end{aligned}
\end{equation}
For a database of time series with varying lengths, the time series anomaly score is computed by normalizing with respective lengths. 



\section{Experiments}\label{experiments}
In this section, we evaluate the performance of the LAD algorithm on multi-aspect datasets. The following experiments have been conducted to study the model:
\begin{enumerate}
    \item Anomaly Detection Performance: LAD's ability to detect real-world anomalies as compared to state-of-the-art anomaly detection models is evaluated using the ground truth labels.
    \item Handling Large Data: Scalability of the LAD model on large datasets (high observation count or high dimensionality) are studied. 
    \item Speed: The computation and execution times of different algorithms are studied and evaluated. 
    \item COVID-19 Time Series Data: We study the performance of LAD model on multiple multivariate time series datasets to identify anomalous instances within each time step as well anomalous time series amongst many.
\end{enumerate}

\subsection{Datasets}
We consider a variety of publicly available benchmark data sets from Outlier Detection DataSets /ODDS~\citep{Rayana:2016} (See Tables ~\ref{tab:datadescription}) for the experimental evaluation. For the time series data, we use COVID-19 deaths and confirmed cases for US counties from John Hopkins COIVD-19 Data Repository \citep{dong2020interactive}. 

\begin{table}[htb]
  \centering
      \begin{tabular}{|l|c|c|c|}
        \toprule
        Name & $N$ & $d$ & $a$ \\
        \midrule
            HTTP  &  567498  &  3  &  0.39\% \\ 
            MNIST  &  7603  &  100  &  9.207\% \\ 
            Arrhythmia  &  452  &  274  &  14.602\% \\ 
            Shuttle  &  49097  &  9  &  7.151\% \\ 
            Letter  &  1600  &  32  &  6.25\% \\ 
            Musk  &  3062  &  166  &  3.168\% \\ 
            Optdigits  &  5216  &  64  &  2.876\% \\ 
            Satellite Image  &  6435  &  36  &  31.639\% \\ 
            Speech  &  3686  &  400  &  1.655\% \\ 
            SMTP  &  95156  &  3  &  0.032\% \\ 
            Satellite Image-2  &  5803  &  36  &  1.224\% \\ 
            Forest Cover  &  286048  &  10  &  0.96\% \\ 
            KDD99  &  620098  &  29  &  29    0.17\% \\ 
      \bottomrule
        \end{tabular}
      \label{tab:datadescriptionlarge}
    \caption{High Dimensional and Large Sample Datasets: Description of the benchmark data sets used for evaluation of the anomaly detection capabilities of the proposed model. $N$ - number of instances, $d$ - number of attributes and $a$ - fraction of known anomalies in the data set.}
    \label{tab:datadescription}
\end{table}

\subsection{Baseline Methods and Parameter Initialization}
As described in Section \ref{methodology}, LAD falls under unsupervised learning regime targeted for high dimensional data, we do not compare with supervised algorithms. For this we consider {\em Elliptic Envelope} (EE)~\cite{rousseeuw1999fast}, {\em Isolation Forest} (I-Forest)~\cite{liu2012isolation}\footnote{The I-Forest model returns both anomaly scores and anomaly labels. As classification model outperforms its score based counterpart on above discussed datasets, we only present results on the classification model.}, {\em local outlier factor} (LOF)~\cite{Breunig:2000},  and {\em Concentration Free Outlier Factor} CFOF \cite{angiulli2020cfof}. The CFOF and LOF models assign an anomaly score for each data instance, while the rest of the methods provide an anomaly label.
As above mentioned methods have one or more user-defined parameters, we investigated a range of values for each parameter, and report the best results. For Isolation Forest, Elliptic Envelope and CFOF, the contamination value is set to the true proportion of anomalies in the dataset.

The LAD model relies on a threshold value to classify observations with scores the value as strictly anomalous. Though this value is iteratively updated, an initial value is required by the algorithm. In this paper, the initial threshold value for the experiment is set to 0.95 for all datasets.


All the methods for anomaly detection benchmark datasets are implemented in Python and all experiments were conducted on a 2.7 GHz Quad-Core Intel Core i7 processor with a 16 GB RAM.

\subsection{Evaluation Metrics}
As LAD is an score based algorithm, we study the ROC curves by comparing the True Positive Rate (TPR) and False Positive Rate (FPR), across various thresholds. The final ROC-AUC (Area under the ROC curve) is reported for evaluation. For time series anomaly detection, we present the final outliers and study their deviations from normal baselines under different model settings.


\subsection{Anomaly Detection Performance}
Table \ref{tab:auc_large} shows the performance of LOF,  I-Forest,  EE, CFOF and LAD on anomaly detection benchmark datasets.  Due to relatively large run-time\footnote{The CFOF model is computationally expensive relative to the rest of the algorithms. As it is aimed to study high-dimensional data, only results on datasets with <10k observations are presented.}, CFOF results are shown for datasets with samples less than 10k. For all the listed algorithms, results for best parameter settings are reported. The proposed LAD model outperforms other methods on most data sets. 
For larger and high-dimensional datasets, it can be seen from Table \ref{tab:auc_large} that the LAD model outperforms all the models in most settings.\footnote{The lowest AUC values for the LAD model are observed for Speech and Optdigits data where multiple true clusters are noted.} 

\begin{table}
    \caption{Comparing LAD with existing anomaly detection algorithms for large/ high dimensional datasets using ROC-AUC as the evaluation metric.}
    \label{tab:auc_large}
  \begin{tabular}{|l|c|c|c|c|c|}
    \toprule
    Data & LOF & I-Forest & EE        & CFOF     & LAD \\
    \midrule
    SHUTTLE & 0.52 & 0.98 & 0.96     &   -        & 0.99 \\
    SATIMAGE-2 & 0.57 & 0.95 & 0.96  &  0.70        & 0.99 \\
    SATIMAGE & 0.51 & 0.64 & 0.65   &   0.55        & 0.6 \\
    KDD99 & 0.51 & 0.85 & 0.54       &      -     & 1.0 \\
    ARRHYTHMIA & 0.61 & 0.67 & 0.7   &    0.56       & 0.71 \\
    OPTDIGITS & 0.51 & 0.52 & 0.45   &    0.49       & 0.48 \\
    LETTER & 0.54 & 0.54 & 0.6       &    0.90       & 0.6 \\
    MUSK & 0.5 & 0.96 & 0.96         &   0.49        & 0.96 \\
    HTTP & 0.47 & 0.95 & 0.95        &      -     & 1.0 \\
    MNIST & 0.5 & 0.61 & 0.65        &  0.75     & 0.87 \\
    COVER & 0.51 & 0.63 & 0.52       &      -     & 0.96 \\
    SMTP & 0.84 & 0.83 & 0.83        &     -      & 0.82 \\
    SPEECH & 0.5 & 0.53 & 0.51       &    0.47  & 0.47 \\
  \bottomrule
\end{tabular}
\end{table}

\begin{figure}
    \centering
    \includegraphics[width=0.45\textwidth]{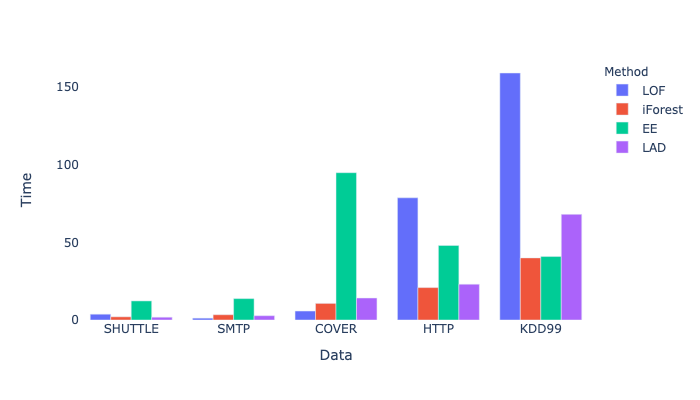}
    \caption{Computation time for large datasets}
    \label{fig:comp_time_long}
\end{figure}

\begin{figure}
    \centering
    \includegraphics[width=0.45\textwidth]{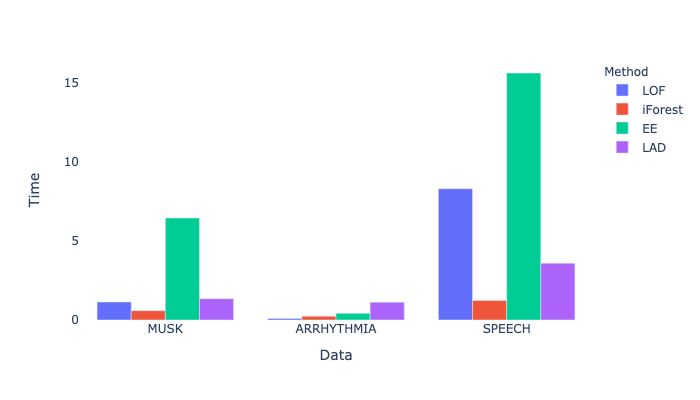}
    \caption{Computation time for high dimensional datasets}
    \label{fig:comp_time_wide}
\end{figure}

To study the LAD model's computational effectiveness, we study the computation time and scaling of LAD model on large and high dimensional datasets. We consider datasets with more than 10k observations or over 100 features for our analysis. Figures \ref{fig:comp_time_long} and \ref{fig:comp_time_wide} show the computation time in seconds for benchmark datasets. It can be seen that the LAD model is relatively low computation time second only to Isolation Forest in most datasets. In fact, the computation time is more stable for our model as opposed to others in high dimensional datasets. 

\begin{figure}
    \centering
    \includegraphics[width=0.45\textwidth]{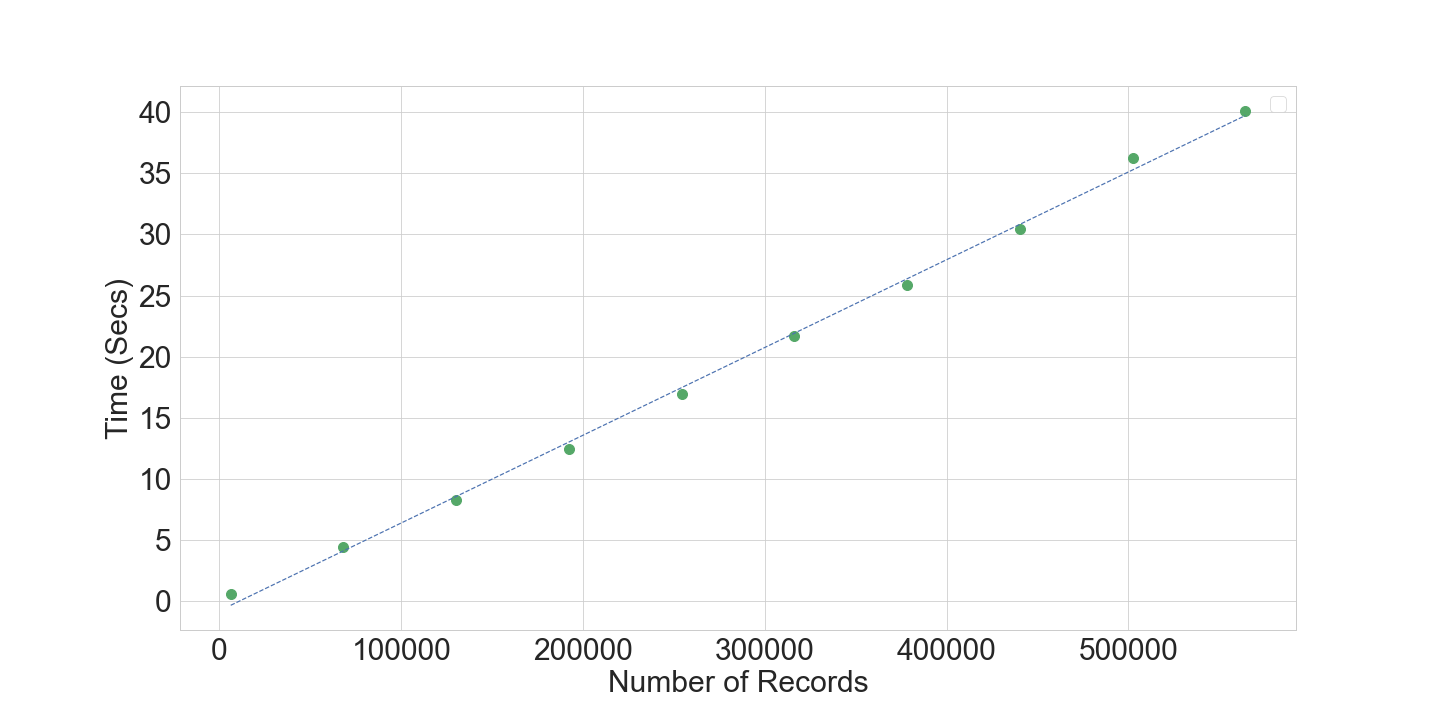}
    \caption{LAD scales linearly with the number of records for KDD-99 data}
    \label{fig:scale_long}
\end{figure}

\begin{figure}
    \centering
    \includegraphics[width=0.45\textwidth]{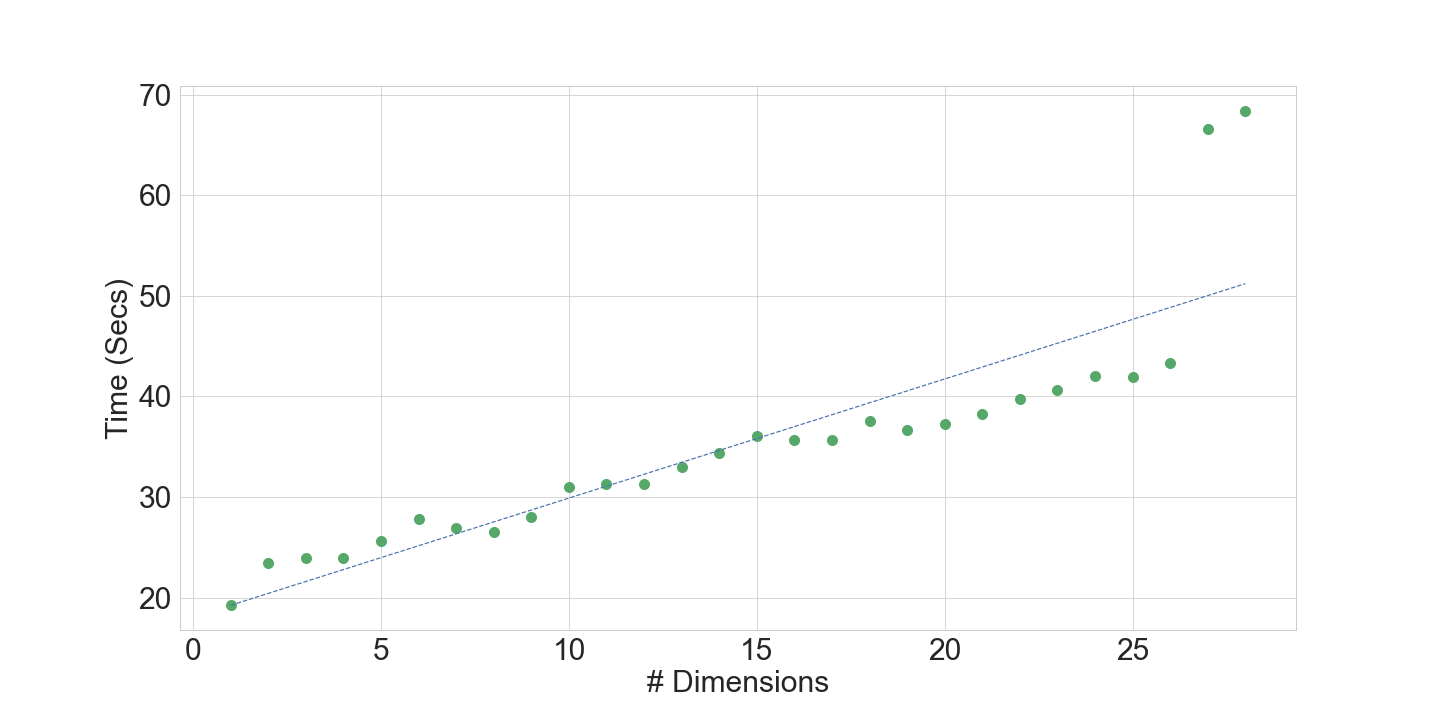}
    \caption{LAD scales linearly with the number of dimensions in KDD-99 data.}
    \label{fig:scale_wide}
\end{figure}

Figure \ref{fig:scale_long} shows the scalability of LAD with respect to the number of records in the data. We plot the time needed to run on the first k records of the KDD-99 dataset. Each record has 29 dimensions. Figure \ref{fig:scale_wide} shows the scalability of LAD with respect to the number of dimensions (linear-scale). We plot the time needed to run on the first 1, 2, ..., 29 dimensions of the KDD-99 dataset.
The results confirm the linear scalability of LAD with number of records  as well as number of dimensions.

\subsection{Anomaly Detection in Time Series Data}
This section presents the results of LAD model on COVID-19 time series data at the US county level. Multiple settings were used to understand the data:
\begin{enumerate}
    \item Deaths and confirmed case trends were considered for analysis
    \item Daily New vs Total Counts: Both total cases as well daily new cases were analyzed for anomaly detection.
    \item Complete history vs One Time Step: Two versions of the model were studied where data from previous time steps were and were not considered. By this, we tried to distinguish the impact of the history of the time series on identifying anomalous trends.
    \item Univariate vs Multivariate Time Series data: To further understand the LAD model, the deaths and case trends were studied individually as a univariate time series  as well as collectively in a multivariate time series data setting. 
    \item Time Series of Uniform vs Varying Lengths: Finally, all the above analyses were conducted on time series data with varying lengths. Here, for each county level time series, the time of first event was considered as initial time step to objectively study the relative temporal changes in trends. 
\end{enumerate}
To bring all the counts to a baseline, the total counts in each time series were scaled to the respective county population. Missing information was replaced with zeros and counties with population less than 50k were eliminated from the study.

\subsection{Discoveries}
\paragraph{Complete history vs One Time Step}
\begin{figure}
     \centering
     \begin{subfigure}[b]{0.49\textwidth}
         \centering
         \includegraphics[width=\textwidth]{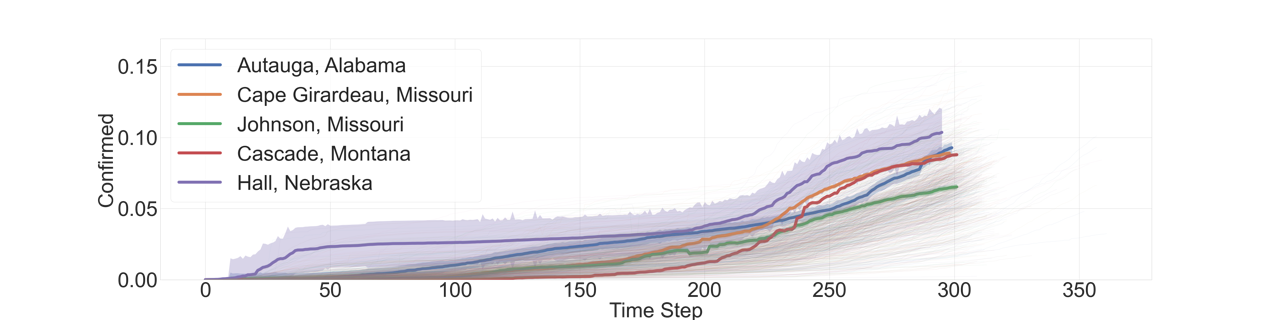}
         \caption{Total Confirmed, Full History}
     \label{fig:uneven_c_total_full_hist_combined}
     \end{subfigure}
     \begin{subfigure}[b]{0.49\textwidth}
         \centering
         \includegraphics[width=\textwidth]{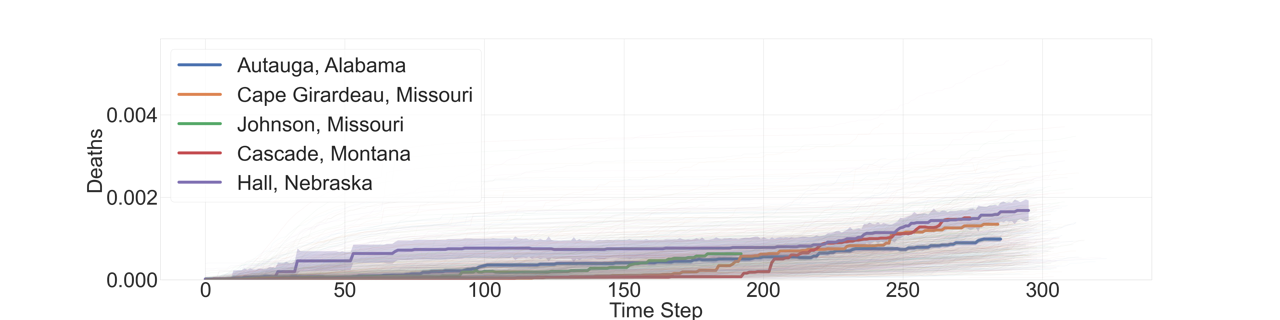}
         \caption{Total Deaths, Full History}
     \label{fig:uneven_d_total_full_hist_combined}
     \end{subfigure}
     \begin{subfigure}[b]{0.49\textwidth}
         \centering
         \includegraphics[width=\textwidth]{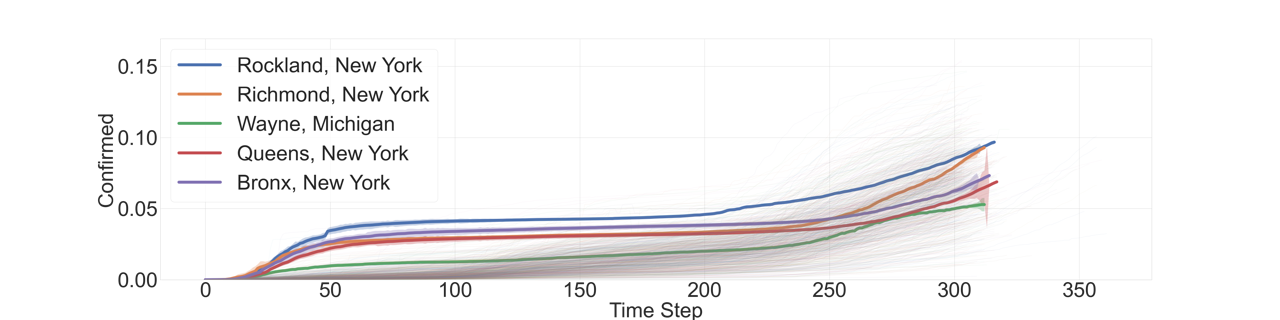}
         \caption{Total Confirmed, One Time Step}
     \label{fig:uneven_c_total_one_timestep_combined}
     \end{subfigure}
     \begin{subfigure}[b]{0.49\textwidth}
         \centering
         \includegraphics[width=\textwidth]{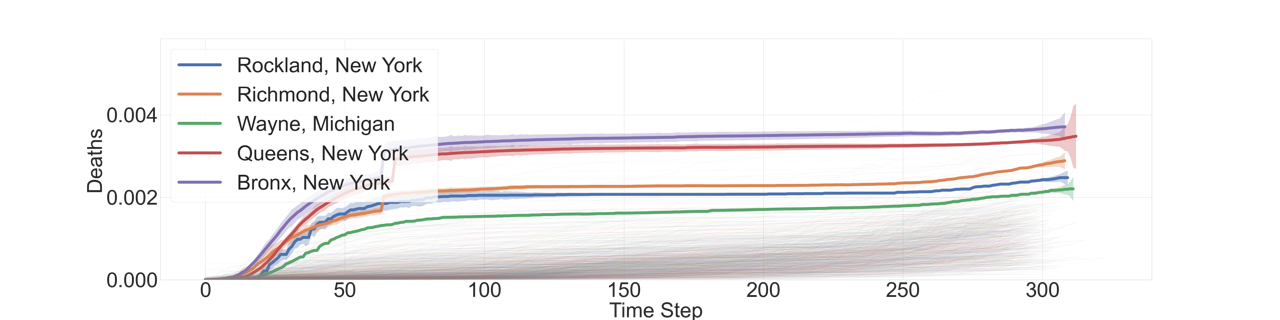}
         \caption{Total Deaths, One Time Step}
     \label{fig:uneven_d_total_one_timestep_combined}
     \end{subfigure}
     \caption{Top 5 Counties with Anomalous Trends : Varying lengths, Total Counts, Multivariate Time Series}
     \label{fig:uneven_total_combined}
\end{figure}

\begin{figure}
     \centering
    \begin{subfigure}[b]{0.49\textwidth}
         \centering
         \includegraphics[width=\textwidth]{covid_images/Confirmed_total_full_history_combined_top_10_per_capita.png}
         \caption{Total Confirmed, Full History}
     \label{fig:c_total_full_hist_combined}
     \end{subfigure}
     \begin{subfigure}[b]{0.49\textwidth}
         \centering
         \includegraphics[width=\textwidth]{covid_images/Deaths_total_full_history_combined_top_10_per_capita.png}
         \caption{Total Deaths, Full History}
     \label{fig:d_total_full_hist_combined}
     \end{subfigure}
     \begin{subfigure}[b]{0.49\textwidth}
         \centering
         \includegraphics[width=\textwidth]{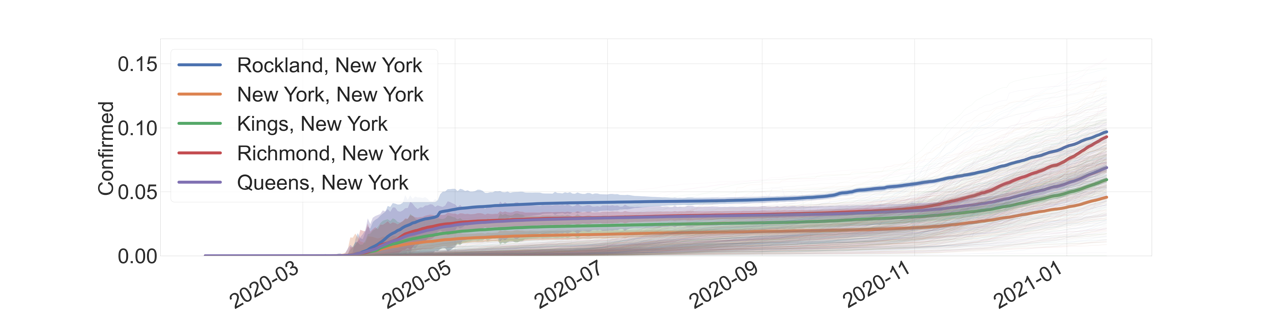}
         \caption{Total Confirmed, One Time Step}
     \label{fig:c_total_one_timestep_combined}
     \end{subfigure}
     \begin{subfigure}[b]{0.49\textwidth}
         \centering
         \includegraphics[width=\textwidth]{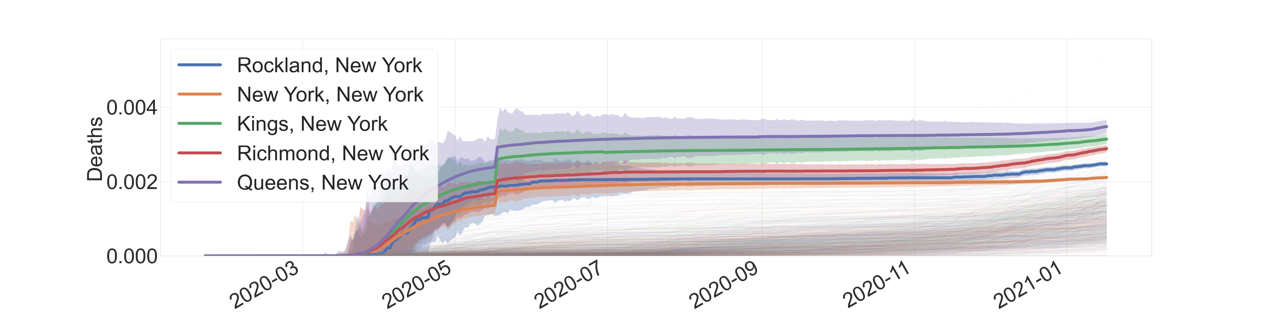}
         \caption{Total Deaths, One Time Step}
     \label{fig:d_total_one_timestep_combined}
     \end{subfigure}
     \caption{Top 5 Counties with Anomalous Trends : Uniform lengths, Total Counts, Multivariate Time Series}
     \label{fig:total_combined}
\end{figure}

The full history setting considers the complete history of the time series and is aimed to capture most deviant trends over time. The one time step (or any smaller window) setting is more suitable to study deviations within the specific window. As we target long term deviating trends, the one time step setting returns trends that have stayed most deviant throughout the entire time range. This can be seen in Figures \ref{fig:uneven_total_combined} and \ref{fig:total_combined} where the one time step setting returns trends that have stayed deviant almost throughout the duration while the full history setting is able to capture significantly wider deviations. For instance, counties like Grand Forks (ND), Burleigh (ND) and Miami-Dale (FL), that had massive outbreaks at later stages\footnote{  \url{https://www.bloomberg.com/news/articles/2020-09-29/north-dakota-s-outbreak-is-as-bad-as-florida-arizona-in-july}} were not captured as anomalous in the one time step model as seen in Figure \ref{fig:c_total_full_hist_combined} and \ref{fig:d_total_full_hist_combined}. Similarly, Hall, Nebraska, which has see a deviation in trend due to an outbreak in meat packing facility in late April 2020, was captured as anomalous trend by the full history model in Figure \ref{fig:uneven_c_total_full_hist_combined} and \ref{fig:uneven_d_total_full_hist_combined}.


\begin{figure}
     \centering
     \begin{subfigure}[b]{0.49\textwidth}
         \centering
         \includegraphics[width=\textwidth]{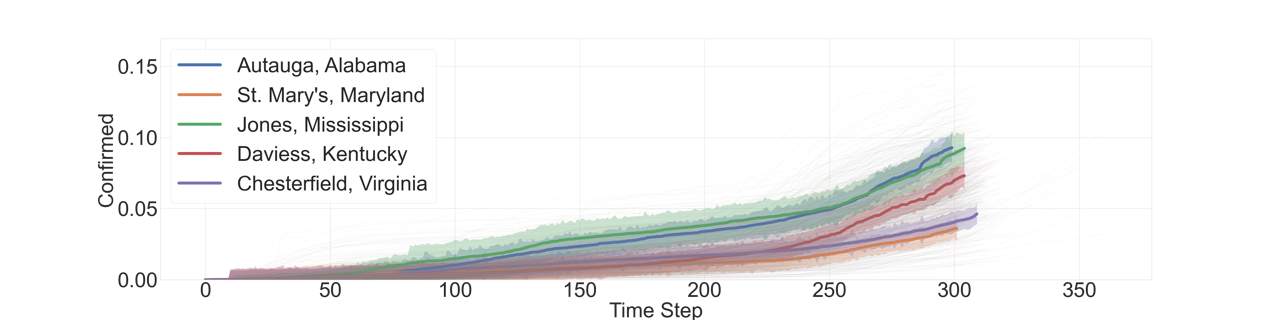}
         \caption{New Confirmed, Full History}
     \label{fig:uneven_c_new_full_hist_combined}
     \end{subfigure}
     \begin{subfigure}[b]{0.49\textwidth}
         \centering
         \includegraphics[width=\textwidth]{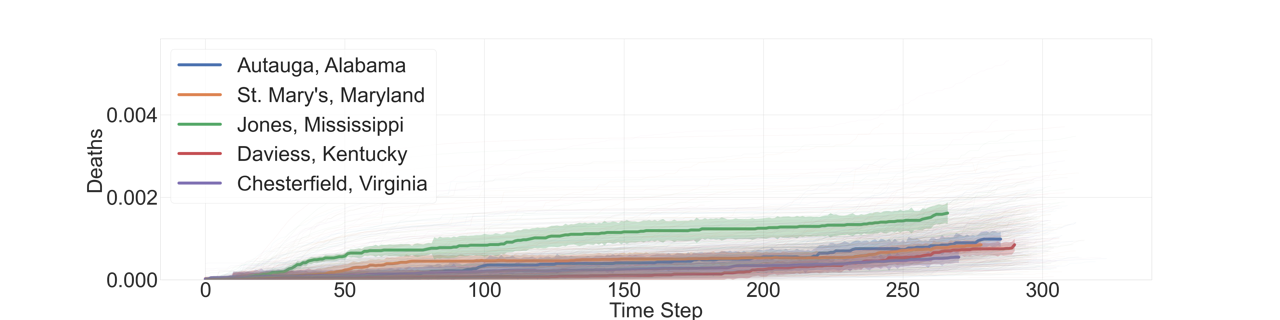}
         \caption{New Deaths, Full History}
     \label{fig:uneven_d_new_full_hist_combined}
     \end{subfigure}
     \caption{Top 5 Counties with Anomalous Trends : Varying lengths, Daily New Counts, Multivariate Time Series}
     \label{fig:uneven_new_combined}
\end{figure}

\begin{figure}
     \centering
     \begin{subfigure}[b]{0.49\textwidth}
         \centering
         \includegraphics[width=\textwidth]{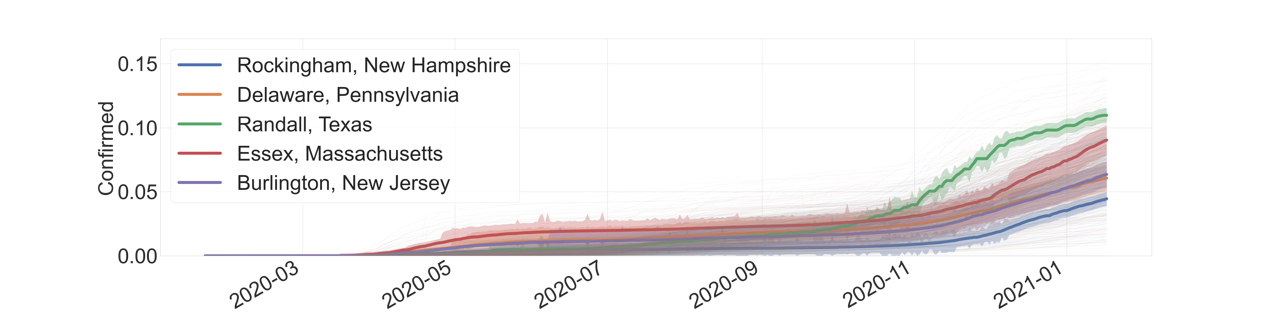}
         \caption{New Confirmed, Full History}
     \label{fig:c_new_full_hist_combined}
     \end{subfigure}
     \begin{subfigure}[b]{0.49\textwidth}
         \centering
         \includegraphics[width=\textwidth]{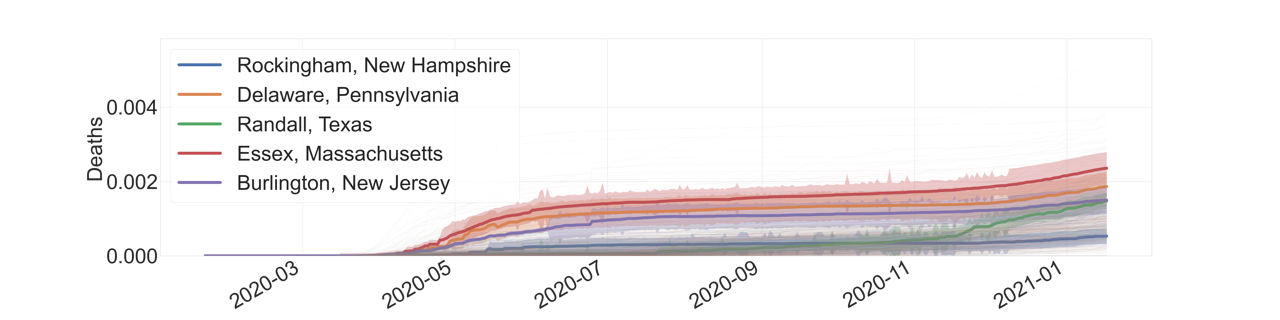}
         \caption{New Deaths, Full History}
     \label{fig:d_new_full_hist_combined}
     \end{subfigure}
     \caption{Top 5 Counties with Anomalous Trends : Uniform lengths, Daily New Counts, Multivariate Time Series}
     \label{fig:new_combined}
\end{figure}

\paragraph{Univariate vs Multivariate Time series}
In Figures \ref{fig:uneven_total_combined}, \ref{fig:total_combined}, \ref{fig:uneven_new_combined} and \ref{fig:new_combined}
we see the anomalous trends in multivariate time series, where total confirmed cases and deaths were collectively evaluated for anomaly detection. For instance, despite the near-normal trends in deaths cases, Hall (NE)\footnote{ \url{https://www.omaha.com/news/state_and_regional/237-coronavirus-cases-tied-to-jbs-beef-plant-in-grand-island-disease-specialists-are-touring/article_2894db56-913a-5c61-a065-6860a8ae50ad.html}} in Figures \ref{fig:uneven_c_total_full_hist_combined}- \ref{fig:uneven_d_total_full_hist_combined}, and Randal (TX) in Figures \ref{fig:c_new_full_hist_combined}-\ref{fig:d_new_full_hist_combined} were identified anomalous due to their the deviant confirmed case trends which significantly contributed to the anomaly scores. This setting enables identification of time-series with at least one deviating feature. 

Similarly, in Figures \ref{fig:uneven_c_total_one_timestep_combined} and \ref{fig:uneven_d_total_one_timestep_combined}, Wayne, Michigan along with Rockland, Richmond, Queens and Bronx in NY have been identified as anomalous. In particular, Michigan was seen to have 3rd highest deaths after NY and NJ in the early stages of the pandemic with Detroit metro-area contributing to most cases\footnote{ \url{https://www.npr.org/sections/coronavirus-live-updates/2020/03/31/824738996/after-surge-in-cases-michigan-now-3rd-in-country-for-coronavirus-deaths}}. Though Wayne county has near normal trend in total confirmed cases where as the total deaths trend has deviated significantly.  

\begin{figure}
     \centering
     \begin{subfigure}[b]{0.49\textwidth}
         \centering
         \includegraphics[width=\textwidth]{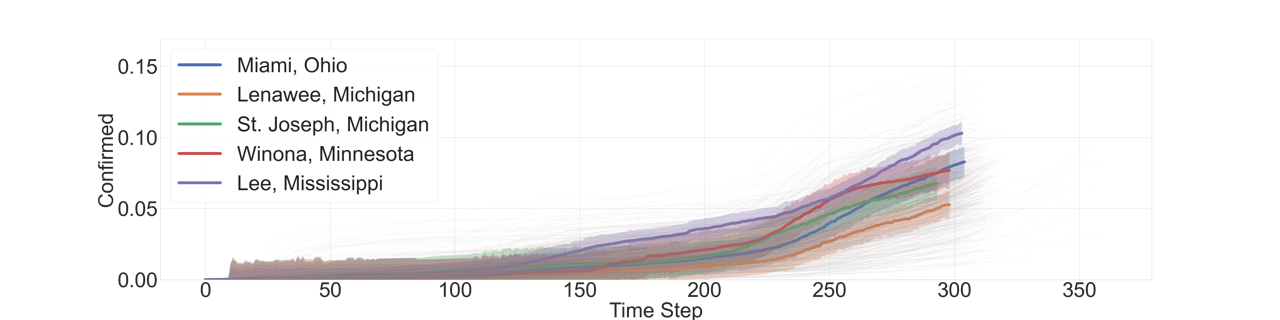}
         \caption{Total Confirmed, Full History}
     \label{fig:uneven_c_total_full_hist}
     \end{subfigure}
     \begin{subfigure}[b]{0.49\textwidth}
         \centering
         \includegraphics[width=\textwidth]{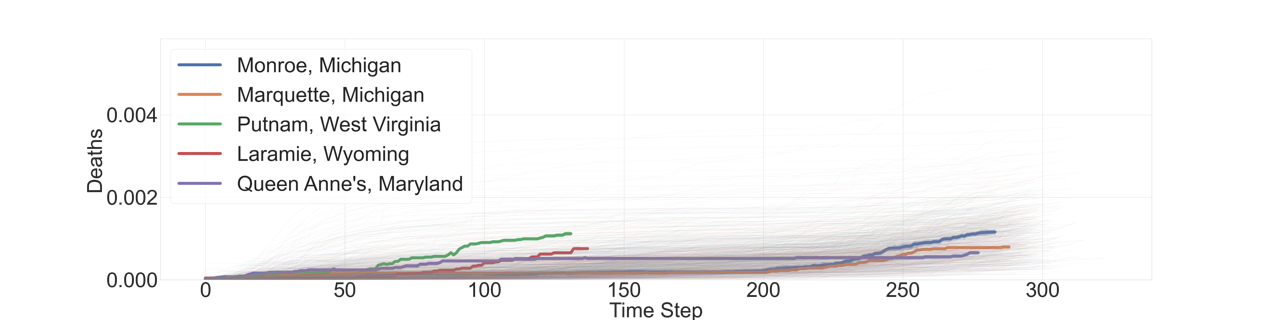}
         \caption{Total Deaths, Full History}
     \label{fig:uneven_d_total_full_hist}
     \end{subfigure}
     \caption{Top 5 Counties with Anomalous Trends: Varying lengths, Total counts}
     \label{fig:uneven_total}
\end{figure}

\begin{figure}
     \centering
     \begin{subfigure}[b]{0.49\textwidth}
         \centering
         \includegraphics[width=\textwidth]{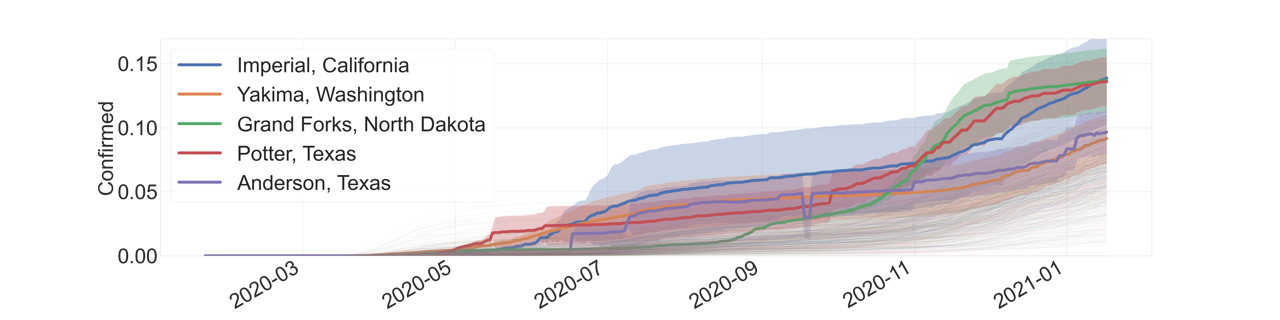}
         \caption{Total Confirmed, Full History}
     \label{fig:c_total_full_hist}
     \end{subfigure}
     \begin{subfigure}[b]{0.49\textwidth}
         \centering
         \includegraphics[width=\textwidth]{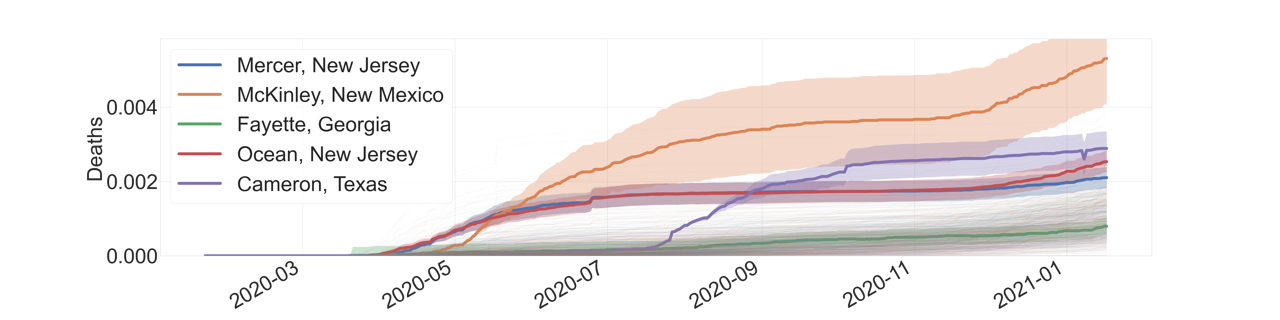}
         \caption{Total Deaths, Full History}
     \label{fig:d_total_full_hist}
     \end{subfigure}
     \caption{Top 5 Counties with Anomalous Trends: Uniform lengths, Total counts}
     \label{fig:total}
\end{figure}

\begin{figure}
     \centering
     \begin{subfigure}[b]{0.49\textwidth}
         \centering
         \includegraphics[width=\textwidth]{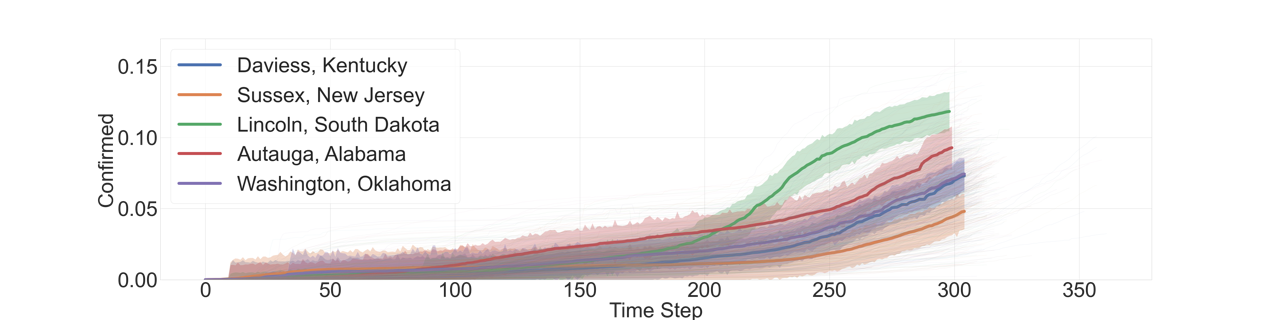}
         \caption{New Confirmed,Full History}
     \label{fig:uneven_c_new_full_hist}
     \end{subfigure}
     \begin{subfigure}[b]{0.49\textwidth}
         \centering
         \includegraphics[width=\textwidth]{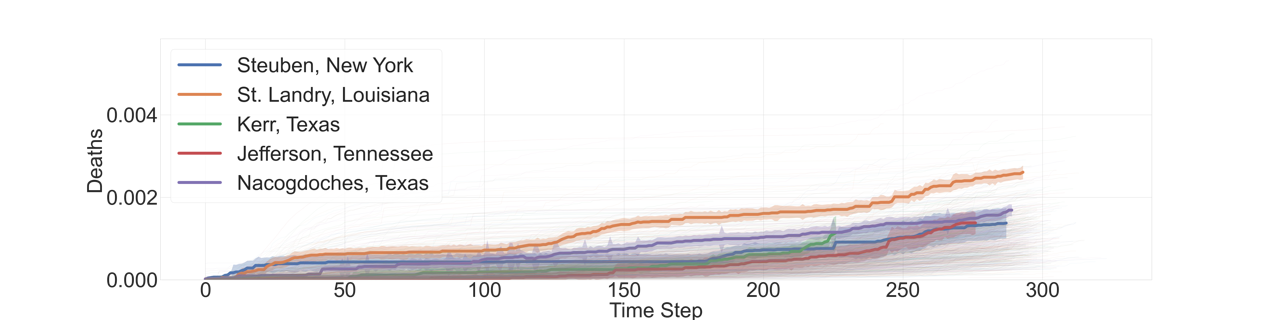}
         \caption{New Deaths, Full History}
     \label{fig:uneven_d_new_full_hist}
     \end{subfigure}
     \caption{Top 5 Counties with Anomalous Trends: Varying lengths, Daily New Counts}
     \label{fig:uneven_new}
\end{figure}

\begin{figure}
     \centering
     \begin{subfigure}[b]{0.49\textwidth}
         \centering
         \includegraphics[width=\textwidth]{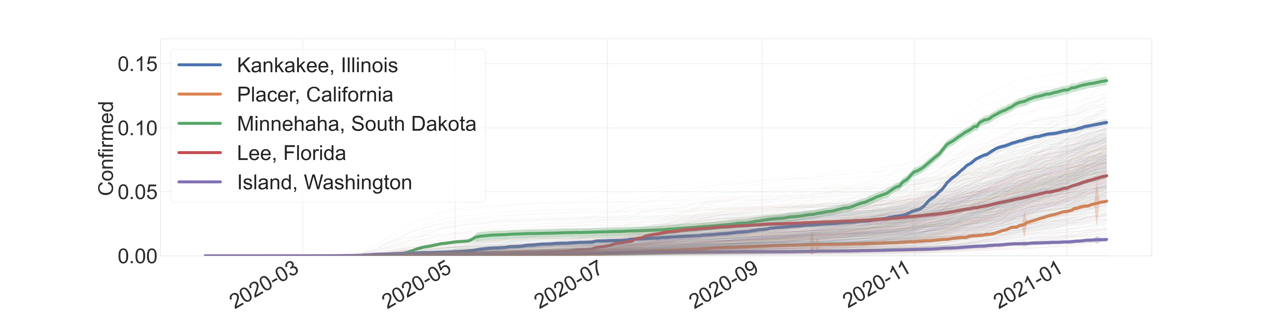}
         \caption{New Confirmed,Full History}
     \label{fig:c_new_full_hist}
     \end{subfigure}
     \begin{subfigure}[b]{0.49\textwidth}
         \centering
         \includegraphics[width=\textwidth]{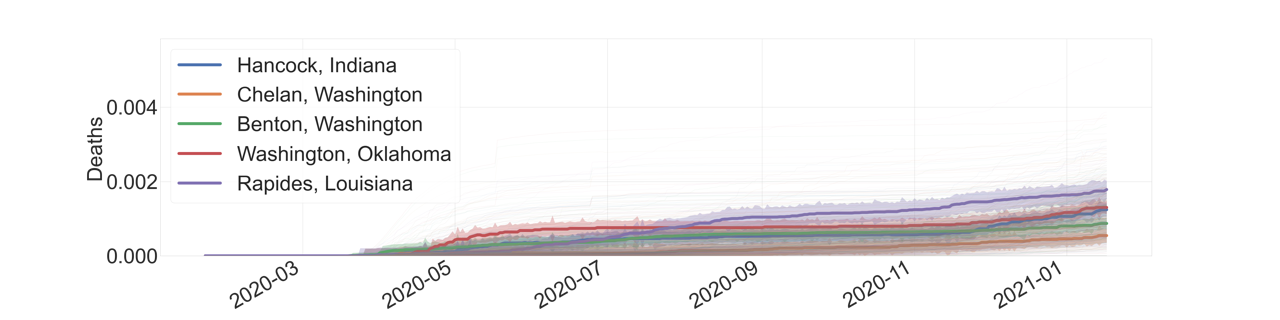}
         \caption{New Deaths, Full History}
     \label{fig:d_new_full_hist}
     \end{subfigure}
     \caption{Top 5 Counties with Anomalous Trends: Uniform lengths, Daily New Counts}
     \label{fig:new}
\end{figure}

\paragraph{Daily New vs Total Counts}
Figures \ref{fig:total_combined} and \ref{fig:new_combined}, show anomalous trends in multivariate time series for total and daily new counts respectively. It can be seen that the anomaly score is erratic for multivariate time series on new case counts. This is due to the fact that the data for new case and death counts is more erratic leading to fluctuating normal average as well as non-smooth anomaly scores. 

The LAD model on the daily new counts data was able to capture the escalation in Greater Boston area, Essex, Massachusetts in Figure \ref{fig:c_new_full_hist_combined} and \ref{fig:d_new_full_hist_combined} during March 2020. Though the total trends seem to be normal, the multiple anomalous daily trends led to their high anomaly scores. Similar patterns led to identification of Lincoln (SD) and Minnehaha (SD) in Figures  \ref{fig:uneven_c_new_full_hist} and \ref{fig:c_new_full_hist} respectively where a subsequent spike occurred after August 2020\footnote{\url{https://www.usatoday.com/story/news/nation/2020/08/07/sturgis-motorcycle-rally-what-know-masks-attendance-rules/3321223001/}}. 

\paragraph{Uniform Length vs Varying Length Time Series}
The US county cases and deaths data consists of time series of uniform lengths. However, not all counties have events recorded in the early stages. Thus, studying the non-synchronized database creates a bias against counties with early reported cases. This can be seen in Figures \ref{fig:uneven_total_combined}  where counties like Wayne, Michigan are flagged anomalous despite starting after many counties in NY and NJ unlike in Figures \ref{fig:total_combined} which reports counties in NY with an early start\footnote{ \url{https://www.npr.org/sections/coronavirus-live-updates/2020/03/31/824738996/after-surge-in-cases-michigan-now-3rd-in-country-for-coronavirus-deaths}}. Similarly, Putnam (WV) and Laramie (WY) are found anomalous in Figure \ref{fig:uneven_d_total_full_hist} where the recently evolved death trends show signs of significant divergence. On the other hand, Potter (TX) and Anderson (TX) have been identified anomalous in Figures \ref{fig:c_total_full_hist} due to early increase in June 2020.

\section{Conclusion}\label{conclusion}
In this paper, we propose LAD, a novel scoring algorithm for anomaly detection in large/high-dimensional data. The algorithm successfully handles high dimensions by implementing large deviation theory. Our contributions include reestablishing the advantages of large deviations theory to large and high dimensional datasets. We also present an online extension of the model that is aimed to identify anomalous time series in a multivariate time series data. The model shows vast potential in scalability and performance against baseline methods. The online LAD returns a temporally evolving score for each time series that allows us to study the deviations in trends relative to the complete time series database. 

A potential extension to the model could include anomalous event detection for each individual time series. Another possible future work could be extending the model to enable anomaly detection in multi-modal datasets. Additionally, the online LAD model could be enhanced to use temporally weighted scores prioritizing recent events. 



\bibliographystyle{ACM-Reference-Format}
\bibliography{main.bib}


\begin{thebibliography}{31}


\ifx \showCODEN    \undefined \def \showCODEN     #1{\unskip}     \fi
\ifx \showDOI      \undefined \def \showDOI       #1{#1}\fi
\ifx \showISBNx    \undefined \def \showISBNx     #1{\unskip}     \fi
\ifx \showISBNxiii \undefined \def \showISBNxiii  #1{\unskip}     \fi
\ifx \showISSN     \undefined \def \showISSN      #1{\unskip}     \fi
\ifx \showLCCN     \undefined \def \showLCCN      #1{\unskip}     \fi
\ifx \shownote     \undefined \def \shownote      #1{#1}          \fi
\ifx \showarticletitle \undefined \def \showarticletitle #1{#1}   \fi
\ifx \showURL      \undefined \def \showURL       {\relax}        \fi
\providecommand\bibfield[2]{#2}
\providecommand\bibinfo[2]{#2}
\providecommand\natexlab[1]{#1}
\providecommand\showeprint[2][]{arXiv:#2}

\bibitem[\protect\citeauthoryear{Aggarwal and Yu}{Aggarwal and Yu}{2001}]%
        {aggarwal2001outlier}
\bibfield{author}{\bibinfo{person}{Charu~C Aggarwal} {and}
  \bibinfo{person}{Philip~S Yu}.} \bibinfo{year}{2001}\natexlab{}.
\newblock \showarticletitle{Outlier detection for high dimensional data}. In
  \bibinfo{booktitle}{\emph{Proceedings of the 2001 ACM SIGMOD international
  conference on Management of data}}. \bibinfo{pages}{37--46}.
\newblock


\bibitem[\protect\citeauthoryear{Angiulli}{Angiulli}{2020}]%
        {angiulli2020cfof}
\bibfield{author}{\bibinfo{person}{Fabrizio Angiulli}.}
  \bibinfo{year}{2020}\natexlab{}.
\newblock \showarticletitle{CFOF: a concentration free measure for anomaly
  detection}.
\newblock \bibinfo{journal}{\emph{ACM Transactions on Knowledge Discovery from
  Data (TKDD)}} \bibinfo{volume}{14}, \bibinfo{number}{1}
  (\bibinfo{year}{2020}), \bibinfo{pages}{1--53}.
\newblock


\bibitem[\protect\citeauthoryear{Angiulli and Pizzuti}{Angiulli and
  Pizzuti}{2002}]%
        {angiulli2002fast}
\bibfield{author}{\bibinfo{person}{Fabrizio Angiulli} {and}
  \bibinfo{person}{Clara Pizzuti}.} \bibinfo{year}{2002}\natexlab{}.
\newblock \showarticletitle{Fast outlier detection in high dimensional spaces}.
  In \bibinfo{booktitle}{\emph{European conference on principles of data mining
  and knowledge discovery}}. Springer, \bibinfo{pages}{15--27}.
\newblock


\bibitem[\protect\citeauthoryear{Beggel, Kausler, Schiegg, Pfeiffer, and
  Bischl}{Beggel et~al\mbox{.}}{2019}]%
        {beggel2019time}
\bibfield{author}{\bibinfo{person}{Laura Beggel}, \bibinfo{person}{Bernhard~X
  Kausler}, \bibinfo{person}{Martin Schiegg}, \bibinfo{person}{Michael
  Pfeiffer}, {and} \bibinfo{person}{Bernd Bischl}.}
  \bibinfo{year}{2019}\natexlab{}.
\newblock \showarticletitle{Time series anomaly detection based on shapelet
  learning}.
\newblock \bibinfo{journal}{\emph{Computational Statistics}}
  \bibinfo{volume}{34}, \bibinfo{number}{3} (\bibinfo{year}{2019}),
  \bibinfo{pages}{945--976}.
\newblock


\bibitem[\protect\citeauthoryear{Benkabou, Benabdeslem, and Canitia}{Benkabou
  et~al\mbox{.}}{2018}]%
        {benkabou2018unsupervised}
\bibfield{author}{\bibinfo{person}{Seif-Eddine Benkabou},
  \bibinfo{person}{Khalid Benabdeslem}, {and} \bibinfo{person}{Bruno Canitia}.}
  \bibinfo{year}{2018}\natexlab{}.
\newblock \showarticletitle{Unsupervised outlier detection for time series by
  entropy and dynamic time warping}.
\newblock \bibinfo{journal}{\emph{Knowledge and Information Systems}}
  \bibinfo{volume}{54}, \bibinfo{number}{2} (\bibinfo{year}{2018}),
  \bibinfo{pages}{463--486}.
\newblock


\bibitem[\protect\citeauthoryear{Boettiger and Hastings}{Boettiger and
  Hastings}{2013}]%
        {boettiger2013no}
\bibfield{author}{\bibinfo{person}{Carl Boettiger} {and} \bibinfo{person}{Alan
  Hastings}.} \bibinfo{year}{2013}\natexlab{}.
\newblock \showarticletitle{No early warning signals for stochastic
  transitions: insights from large deviation theory}.
\newblock \bibinfo{journal}{\emph{Proceedings of the Royal Society B:
  Biological Sciences}} \bibinfo{volume}{280}, \bibinfo{number}{1766}
  (\bibinfo{year}{2013}), \bibinfo{pages}{20131372}.
\newblock


\bibitem[\protect\citeauthoryear{Breunig, Kriegel, Ng, and Sander}{Breunig
  et~al\mbox{.}}{2000}]%
        {Breunig:2000}
\bibfield{author}{\bibinfo{person}{Markus~M. Breunig},
  \bibinfo{person}{Hans-Peter Kriegel}, \bibinfo{person}{Raymond~T. Ng}, {and}
  \bibinfo{person}{J. Sander}.} \bibinfo{year}{2000}\natexlab{}.
\newblock \showarticletitle{LOF: identifying density-based local outliers}. In
  \bibinfo{booktitle}{\emph{Proceedings of 2000 ACM SIGMOD International
  Conference on Management of Data}}. \bibinfo{pages}{93--104}.
\newblock


\bibitem[\protect\citeauthoryear{Ceylan}{Ceylan}{2020}]%
        {ceylan2020estimation}
\bibfield{author}{\bibinfo{person}{Zeynep Ceylan}.}
  \bibinfo{year}{2020}\natexlab{}.
\newblock \showarticletitle{Estimation of COVID-19 prevalence in Italy, Spain,
  and France}.
\newblock \bibinfo{journal}{\emph{Science of The Total Environment}}
  \bibinfo{volume}{729} (\bibinfo{year}{2020}), \bibinfo{pages}{138817}.
\newblock


\bibitem[\protect\citeauthoryear{Chandola, Banerjee, and Kumar}{Chandola
  et~al\mbox{.}}{2009a}]%
        {chandola2009anomaly}
\bibfield{author}{\bibinfo{person}{Varun Chandola}, \bibinfo{person}{Arindam
  Banerjee}, {and} \bibinfo{person}{Vipin Kumar}.}
  \bibinfo{year}{2009}\natexlab{a}.
\newblock \showarticletitle{Anomaly detection: A survey}.
\newblock \bibinfo{journal}{\emph{Comput. Surveys}} \bibinfo{volume}{41},
  \bibinfo{number}{3} (\bibinfo{year}{2009}).
\newblock


\bibitem[\protect\citeauthoryear{Chandola, Cheboli, and Kumar}{Chandola
  et~al\mbox{.}}{2009b}]%
        {Chandola:2009a}
\bibfield{author}{\bibinfo{person}{Varun Chandola}, \bibinfo{person}{Deepthi
  Cheboli}, {and} \bibinfo{person}{Vipin Kumar}.}
  \bibinfo{year}{2009}\natexlab{b}.
\newblock \bibinfo{booktitle}{\emph{Detecting Anomalies in a Timeseries
  Database}}.
\newblock \bibinfo{type}{{T}echnical {R}eport} 09-004.
  \bibinfo{institution}{University of Minnesota, Computer Science Department}.
\newblock


\bibitem[\protect\citeauthoryear{Chandola and Kumar}{Chandola and
  Kumar}{2008}]%
        {Chandola:2008}
\bibfield{author}{\bibinfo{person}{V. Chandola} {and} \bibinfo{person}{V.
  Kumar}.} \bibinfo{year}{2008}\natexlab{}.
\newblock \showarticletitle{A Comparative Evaluation of Anomaly Detection
  Techniques for Sequence Data}. In \bibinfo{booktitle}{\emph{Proceedings of
  International Conference on Data Mining}}. \bibinfo{address}{Pisa, Italy}.
\newblock


\bibitem[\protect\citeauthoryear{Chawla and Gionis}{Chawla and Gionis}{2013}]%
        {chawla2013k}
\bibfield{author}{\bibinfo{person}{Sanjay Chawla} {and}
  \bibinfo{person}{Aristides Gionis}.} \bibinfo{year}{2013}\natexlab{}.
\newblock \showarticletitle{k-means--: A unified approach to clustering and
  outlier detection}. In \bibinfo{booktitle}{\emph{SDM}}.
\newblock


\bibitem[\protect\citeauthoryear{Dematteis, Grafke, and
  Vanden-Eijnden}{Dematteis et~al\mbox{.}}{2018}]%
        {dematteis2018rogue}
\bibfield{author}{\bibinfo{person}{Giovanni Dematteis}, \bibinfo{person}{Tobias
  Grafke}, {and} \bibinfo{person}{Eric Vanden-Eijnden}.}
  \bibinfo{year}{2018}\natexlab{}.
\newblock \showarticletitle{Rogue waves and large deviations in deep sea}.
\newblock \bibinfo{journal}{\emph{Proceedings of the National Academy of
  Sciences}} \bibinfo{volume}{115}, \bibinfo{number}{5} (\bibinfo{year}{2018}),
  \bibinfo{pages}{855--860}.
\newblock


\bibitem[\protect\citeauthoryear{Den~Hollander}{Den~Hollander}{2008}]%
        {den2008large}
\bibfield{author}{\bibinfo{person}{Frank Den~Hollander}.}
  \bibinfo{year}{2008}\natexlab{}.
\newblock \bibinfo{booktitle}{\emph{Large deviations}}.
  Vol.~\bibinfo{volume}{14}.
\newblock \bibinfo{publisher}{American Mathematical Soc.}
\newblock


\bibitem[\protect\citeauthoryear{Dong, Du, and Gardner}{Dong
  et~al\mbox{.}}{2020}]%
        {dong2020interactive}
\bibfield{author}{\bibinfo{person}{Ensheng Dong}, \bibinfo{person}{Hongru Du},
  {and} \bibinfo{person}{Lauren Gardner}.} \bibinfo{year}{2020}\natexlab{}.
\newblock \showarticletitle{An interactive web-based dashboard to track
  COVID-19 in real time}.
\newblock \bibinfo{journal}{\emph{The Lancet infectious diseases}}
  \bibinfo{volume}{20}, \bibinfo{number}{5} (\bibinfo{year}{2020}),
  \bibinfo{pages}{533--534}.
\newblock


\bibitem[\protect\citeauthoryear{Fu}{Fu}{2011}]%
        {fu2011review}
\bibfield{author}{\bibinfo{person}{Tak-chung Fu}.}
  \bibinfo{year}{2011}\natexlab{}.
\newblock \showarticletitle{A review on time series data mining}.
\newblock \bibinfo{journal}{\emph{Engineering Applications of Artificial
  Intelligence}} \bibinfo{volume}{24}, \bibinfo{number}{1}
  (\bibinfo{year}{2011}), \bibinfo{pages}{164--181}.
\newblock


\bibitem[\protect\citeauthoryear{Gupta, Gao, Aggarwal, and Han}{Gupta
  et~al\mbox{.}}{2013}]%
        {gupta2013outlier}
\bibfield{author}{\bibinfo{person}{Manish Gupta}, \bibinfo{person}{Jing Gao},
  \bibinfo{person}{Charu~C Aggarwal}, {and} \bibinfo{person}{Jiawei Han}.}
  \bibinfo{year}{2013}\natexlab{}.
\newblock \showarticletitle{Outlier detection for temporal data: A survey}.
\newblock \bibinfo{journal}{\emph{IEEE Transactions on Knowledge and data
  Engineering}} \bibinfo{volume}{26}, \bibinfo{number}{9}
  (\bibinfo{year}{2013}), \bibinfo{pages}{2250--2267}.
\newblock


\bibitem[\protect\citeauthoryear{Hodge and Austin}{Hodge and Austin}{2004}]%
        {Hodge:2004}
\bibfield{author}{\bibinfo{person}{Victoria Hodge} {and} \bibinfo{person}{Jim
  Austin}.} \bibinfo{year}{2004}\natexlab{}.
\newblock \showarticletitle{A Survey of Outlier Detection Methodologies}.
\newblock \bibinfo{journal}{\emph{Artificial Intelligence Review}}
  \bibinfo{volume}{22}, \bibinfo{number}{2} (\bibinfo{year}{2004}),
  \bibinfo{pages}{85--126}.
\newblock


\bibitem[\protect\citeauthoryear{Liu, Ting, and Zhou}{Liu
  et~al\mbox{.}}{2012}]%
        {liu2012isolation}
\bibfield{author}{\bibinfo{person}{Fei~Tony Liu}, \bibinfo{person}{Kai~Ming
  Ting}, {and} \bibinfo{person}{Zhi-Hua Zhou}.}
  \bibinfo{year}{2012}\natexlab{}.
\newblock \showarticletitle{Isolation-based anomaly detection}.
\newblock \bibinfo{journal}{\emph{ACM Transactions on Knowledge Discovery from
  Data (TKDD)}} \bibinfo{volume}{6}, \bibinfo{number}{1}
  (\bibinfo{year}{2012}), \bibinfo{pages}{1--39}.
\newblock


\bibitem[\protect\citeauthoryear{Maleki, Mahmoudi, Wraith, and Pho}{Maleki
  et~al\mbox{.}}{2020}]%
        {maleki2020time}
\bibfield{author}{\bibinfo{person}{Mohsen Maleki},
  \bibinfo{person}{Mohammad~Reza Mahmoudi}, \bibinfo{person}{Darren Wraith},
  {and} \bibinfo{person}{Kim-Hung Pho}.} \bibinfo{year}{2020}\natexlab{}.
\newblock \showarticletitle{Time series modelling to forecast the confirmed and
  recovered cases of COVID-19}.
\newblock \bibinfo{journal}{\emph{Travel medicine and infectious disease}}
  \bibinfo{volume}{37} (\bibinfo{year}{2020}), \bibinfo{pages}{101742}.
\newblock


\bibitem[\protect\citeauthoryear{Mikosch and Wintenberger}{Mikosch and
  Wintenberger}{2016}]%
        {mikosch2016large}
\bibfield{author}{\bibinfo{person}{Thomas Mikosch} {and}
  \bibinfo{person}{Olivier Wintenberger}.} \bibinfo{year}{2016}\natexlab{}.
\newblock \showarticletitle{A large deviations approach to limit theory for
  heavy-tailed time series}.
\newblock \bibinfo{journal}{\emph{Probability Theory and Related Fields}}
  \bibinfo{volume}{166}, \bibinfo{number}{1} (\bibinfo{year}{2016}),
  \bibinfo{pages}{233--269}.
\newblock


\bibitem[\protect\citeauthoryear{Paschalidis and Smaragdakis}{Paschalidis and
  Smaragdakis}{2008}]%
        {paschalidis2008spatio}
\bibfield{author}{\bibinfo{person}{Ioannis~Ch Paschalidis} {and}
  \bibinfo{person}{Georgios Smaragdakis}.} \bibinfo{year}{2008}\natexlab{}.
\newblock \showarticletitle{Spatio-temporal network anomaly detection by
  assessing deviations of empirical measures}.
\newblock \bibinfo{journal}{\emph{IEEE/ACM Transactions On Networking}}
  \bibinfo{volume}{17}, \bibinfo{number}{3} (\bibinfo{year}{2008}),
  \bibinfo{pages}{685--697}.
\newblock


\bibitem[\protect\citeauthoryear{Ramaswamy, Rastogi, and Shim}{Ramaswamy
  et~al\mbox{.}}{2000}]%
        {Ramaswamy:2000}
\bibfield{author}{\bibinfo{person}{Sridhar Ramaswamy}, \bibinfo{person}{Rajeev
  Rastogi}, {and} \bibinfo{person}{Kyuseok Shim}.}
  \bibinfo{year}{2000}\natexlab{}.
\newblock \showarticletitle{Efficient algorithms for mining outliers from large
  data sets}. In \bibinfo{booktitle}{\emph{Proceedings of the 2000 ACM SIGMOD
  international conference on Management of data}}. \bibinfo{publisher}{ACM
  Press}, \bibinfo{address}{Dallas, Texas, United States},
  \bibinfo{pages}{427--438}.
\newblock
\showISBNx{1-58113-217-4}


\bibitem[\protect\citeauthoryear{Rayana}{Rayana}{2016}]%
        {Rayana:2016}
\bibfield{author}{\bibinfo{person}{Shebuti Rayana}.}
  \bibinfo{year}{2016}\natexlab{}.
\newblock \bibinfo{title}{{ODDS} Library}.
\newblock
\newblock
\urldef\tempurl%
\url{http://odds.cs.stonybrook.edu}
\showURL{%
\tempurl}


\bibitem[\protect\citeauthoryear{Rousseeuw and Driessen}{Rousseeuw and
  Driessen}{1999}]%
        {rousseeuw1999fast}
\bibfield{author}{\bibinfo{person}{Peter~J Rousseeuw} {and}
  \bibinfo{person}{Katrien~Van Driessen}.} \bibinfo{year}{1999}\natexlab{}.
\newblock \showarticletitle{A fast algorithm for the minimum covariance
  determinant estimator}.
\newblock \bibinfo{journal}{\emph{Technometrics}} \bibinfo{volume}{41},
  \bibinfo{number}{3} (\bibinfo{year}{1999}), \bibinfo{pages}{212--223}.
\newblock


\bibitem[\protect\citeauthoryear{Touchette}{Touchette}{2009}]%
        {touchette2009large}
\bibfield{author}{\bibinfo{person}{Hugo Touchette}.}
  \bibinfo{year}{2009}\natexlab{}.
\newblock \showarticletitle{The large deviation approach to statistical
  mechanics}.
\newblock \bibinfo{journal}{\emph{Physics Reports}} \bibinfo{volume}{478},
  \bibinfo{number}{1-3} (\bibinfo{year}{2009}), \bibinfo{pages}{1--69}.
\newblock


\bibitem[\protect\citeauthoryear{Varadhan}{Varadhan}{1984}]%
        {varadhan1984large}
\bibfield{author}{\bibinfo{person}{SR~Srinivasa Varadhan}.}
  \bibinfo{year}{1984}\natexlab{}.
\newblock \bibinfo{booktitle}{\emph{Large deviations and applications}}.
\newblock \bibinfo{publisher}{SIAM}.
\newblock


\bibitem[\protect\citeauthoryear{Varadhan}{Varadhan}{2010}]%
        {varadhan2010large}
\bibfield{author}{\bibinfo{person}{SR~Srinivasa Varadhan}.}
  \bibinfo{year}{2010}\natexlab{}.
\newblock \showarticletitle{Large deviations}. In
  \bibinfo{booktitle}{\emph{Proceedings of the International Congress of
  Mathematicians 2010 (ICM 2010) (In 4 Volumes) Vol. I: Plenary Lectures and
  Ceremonies Vols. II--IV: Invited Lectures}}. World Scientific,
  \bibinfo{pages}{622--639}.
\newblock


\bibitem[\protect\citeauthoryear{Yankov, Keogh, and Rebbapragada}{Yankov
  et~al\mbox{.}}{2008}]%
        {yankov2008disk}
\bibfield{author}{\bibinfo{person}{Dragomir Yankov}, \bibinfo{person}{Eamonn
  Keogh}, {and} \bibinfo{person}{Umaa Rebbapragada}.}
  \bibinfo{year}{2008}\natexlab{}.
\newblock \showarticletitle{Disk aware discord discovery: Finding unusual time
  series in terabyte sized datasets}.
\newblock \bibinfo{journal}{\emph{Knowledge and Information Systems}}
  \bibinfo{volume}{17}, \bibinfo{number}{2} (\bibinfo{year}{2008}),
  \bibinfo{pages}{241--262}.
\newblock


\bibitem[\protect\citeauthoryear{Zeroual, Harrou, Dairi, and Sun}{Zeroual
  et~al\mbox{.}}{2020}]%
        {zeroual2020deep}
\bibfield{author}{\bibinfo{person}{Abdelhafid Zeroual}, \bibinfo{person}{Fouzi
  Harrou}, \bibinfo{person}{Abdelkader Dairi}, {and} \bibinfo{person}{Ying
  Sun}.} \bibinfo{year}{2020}\natexlab{}.
\newblock \showarticletitle{Deep learning methods for forecasting COVID-19
  time-Series data: A Comparative study}.
\newblock \bibinfo{journal}{\emph{Chaos, Solitons \& Fractals}}
  \bibinfo{volume}{140} (\bibinfo{year}{2020}), \bibinfo{pages}{110121}.
\newblock


\bibitem[\protect\citeauthoryear{Zhang, Song, Chen, Feng, Lumezanu, Cheng, Ni,
  Zong, Chen, and Chawla}{Zhang et~al\mbox{.}}{2019}]%
        {zhang2019deep}
\bibfield{author}{\bibinfo{person}{Chuxu Zhang}, \bibinfo{person}{Dongjin
  Song}, \bibinfo{person}{Yuncong Chen}, \bibinfo{person}{Xinyang Feng},
  \bibinfo{person}{Cristian Lumezanu}, \bibinfo{person}{Wei Cheng},
  \bibinfo{person}{Jingchao Ni}, \bibinfo{person}{Bo Zong},
  \bibinfo{person}{Haifeng Chen}, {and} \bibinfo{person}{Nitesh~V Chawla}.}
  \bibinfo{year}{2019}\natexlab{}.
\newblock \showarticletitle{A deep neural network for unsupervised anomaly
  detection and diagnosis in multivariate time series data}. In
  \bibinfo{booktitle}{\emph{Proceedings of the AAAI Conference on Artificial
  Intelligence}}, Vol.~\bibinfo{volume}{33}. \bibinfo{pages}{1409--1416}.
\newblock


\end{thebibliography}
\end{document}